\documentclass[10pt,twocolumn,letterpaper]{article}

\usepackage{iccv}
\usepackage{times}
\usepackage{epsfig}
\usepackage{graphicx}
\usepackage{amsmath}
\usepackage{amssymb}
\usepackage{makecell}
\usepackage{multirow}
\usepackage{caption}
\usepackage{mathtools}
\usepackage{appendix}

\usepackage[pagebackref=true,breaklinks=true,colorlinks,bookmarks=false]{hyperref}

\iccvfinalcopy

\begin{document} 

\title{MINR: Implicit Neural Representations with Masked Image Modelling}
\author{
    Sua Lee\thanks{Equal contribution}\quad
    Joonhun Lee$^*$\quad
    Myungjoo Kang\thanks{Corresponding author}\\
    Seoul National University\\
    {\tt\small\{susan900,niceguy718,mkang\}@snu.ac.kr}
}
\maketitle

\pagestyle{empty}\thispagestyle{empty}

\begin{abstract}
   Self-supervised learning methods like masked autoencoders (MAE) have shown significant promise in learning robust feature representations, particularly in image reconstruction-based pretraining task. However, their performance is often strongly dependent on the masking strategies used during training and can degrade when applied to out-of-distribution data. To address these limitations, we introduce the masked implicit neural representations (MINR) framework that synergizes implicit neural representations with masked image modeling. MINR learns a continuous function to represent images, enabling more robust and generalizable reconstructions irrespective of masking strategies. Our experiments demonstrate that MINR not only outperforms MAE in in-domain scenarios but also in out-of-distribution settings, while reducing model complexity. The versatility of MINR extends to various self-supervised learning applications, confirming its utility as a robust and efficient alternative to existing frameworks.
\end{abstract}

\section{Introduction}
Deep learning methods have rapidly advanced with supervised learning in computer vision, but it struggles with significant performance degradation when tested on the data distributions not observed during the training phase.
This challenge stems from the inherent assumption in supervised learning: the training and test sets are drawn independently and identically from the same underlying data distribution.
However, deep learning models often encounter situations requiring adaptation to unseen data distributions; generalizing effectively across such distributions is crucial for ensuring model robustness.

\begin{figure}[t]
    \begin{center}
        \includegraphics[width=.95\columnwidth]{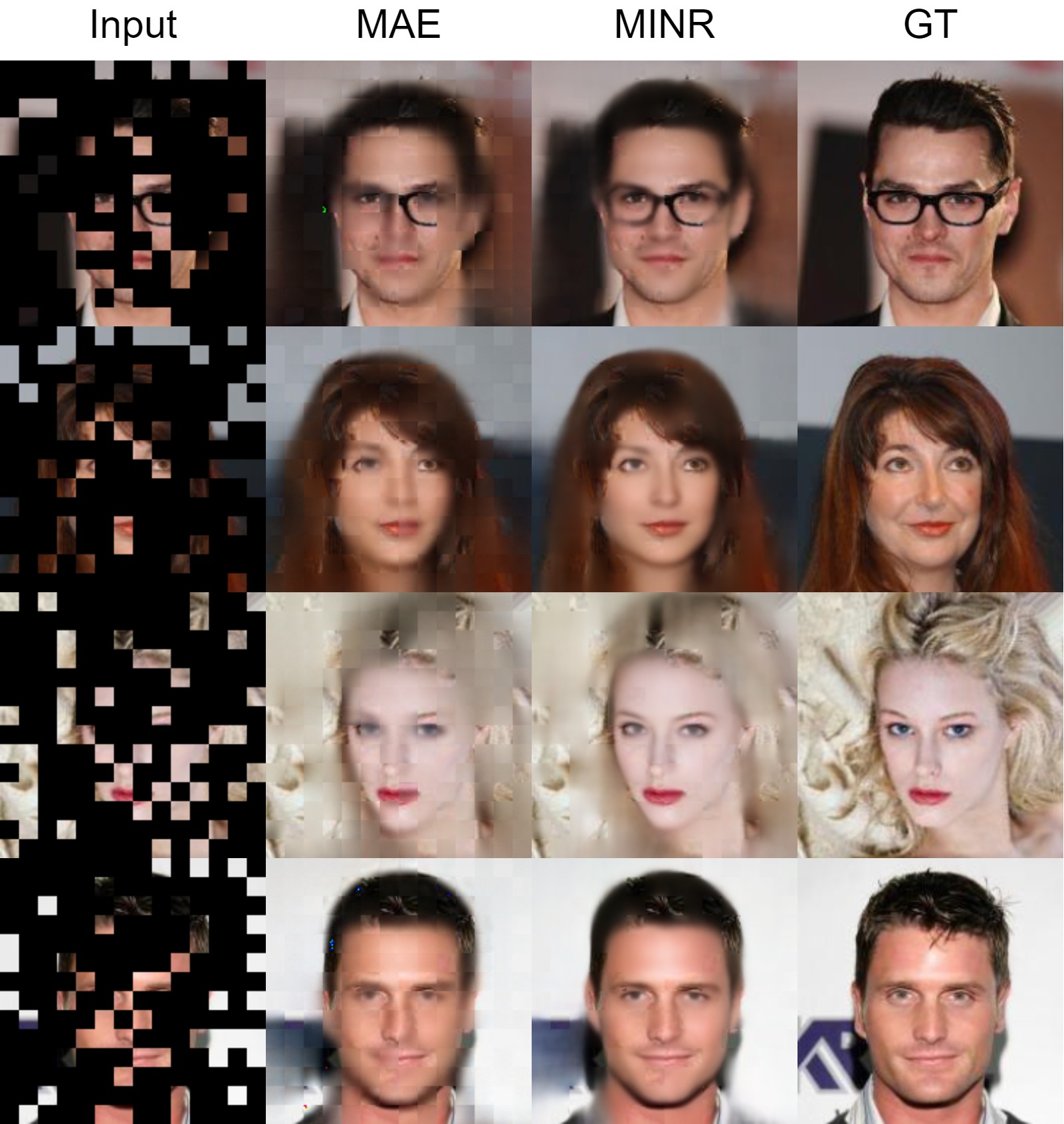}
    \end{center}
    \caption{
        \textbf{Qualitative results of mask reconstruction.}
        For each row, we present the masked image, MAE and MINR reconstructions, and the ground truth, in sequence. 
    }
    \label{fig:result_figure}
\end{figure}

\begin{figure*}[t]
    \begin{center}
        \includegraphics[width=1.7\columnwidth]{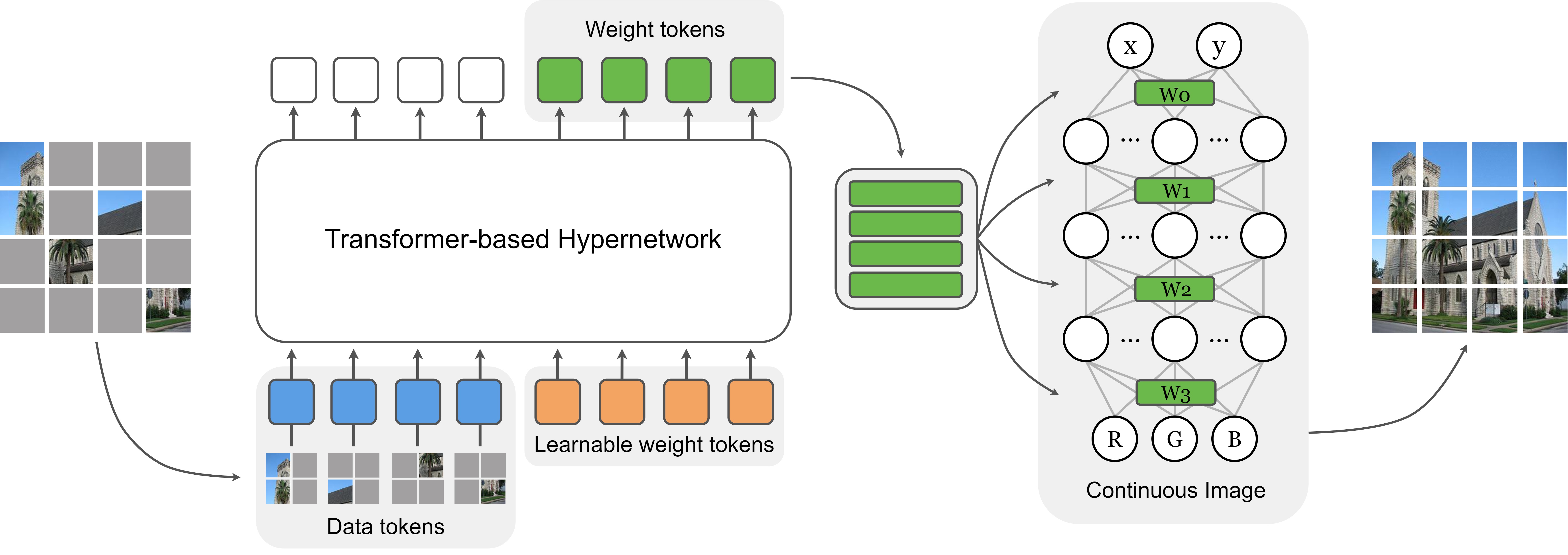}
    \end{center}
    \captionsetup{width=2.0\columnwidth}
    \caption[width=\textwidth]{
        \textbf{A schematic illustration of MINR.}
        During the training phase, a large random subset of image patches are masked out.
        To ensure robustness, we employ a transformer-based hypernetwork predicting weights for an INR.
        TransINR directly maps weight sets \cite{chen2022transformers}, whereas GINR modulates only the second MLP layer as instance-specific, keeping the rest instance-agnostic \cite{kim2023generalizable}.
        The overall framework constructs weights with the masked-out input image and outputs an interpolated one.
    }
    \label{fig:framework overview}
\end{figure*}

To address this, self-supervised learning (SSL) methods have gained attention.
Using pretext tasks, SSL learns robust feature representations without human-annotated labels.
Such feature representations have demonstrated increased robustness and performance in varied tasks domain, including domain generalization scenarios \cite{albuquerque2020improving, hendrycks2019using}.
Notable SSL techniques include masked image modelling (MIM) that enhance representation robustness by deliberately masking images and training models to reconstruct the hidden information.
The effectiveness of MIM has been demonstrated in various downstream tasks, remaining one of the most powerful pretraining methods.

Masked autoencoder (MAE) has been often highlighted for its versatility and success in various tasks such as ConvNeXt V2 in image recognition \cite{woo2023convnext}, DropMAE in video representation \cite{wu2023dropmae}, and PiMAE in 3D object detection \cite{chen2023pimae}.
Although certain frameworks may marginally outperform MAE in some scenarios, MAE's end-to-end approach simplifies the training process by eliminating the need for a separate pretrained model.

However, a notable limitation of MAE is its dependency on masking strategies, such as mask size and area, as evidenced by several studies \cite{kakogeorgiou2022hide,shi2022adversarial,wang2023hard}.
The MAE not only utilizes adjacent patches but also employs explicit information from all visible patches to fill each masked patch \cite{cao2022understand}.
As shown in \cite{kong2023understanding}, the hierarchical extraction of explicit information---to fill the masked patches---plays an instrumental role in shaping the learned representation; it is directly influenced by the amount and quality of information in the visible patches.
Furthermore, MAE computes the loss only on masked patches during training, thus, in testing, employing a masking strategy unseen during the training phase leads to a drastic decline in performance.

In this work, we introduce the \emph{masked implicit neural representations (MINR)} framework that combines implicit neural representations (INRs) with MIM to address the limitations of MAE.
The advantages of MINR include:
\lowercase\expandafter{\romannumeral1}) Leveraging INRs to learn a continuous function less affected by variations of information in visible patches, resulting in performance improvements in both in-domain and out-of-distribution settings;
\lowercase\expandafter{\romannumeral2}) Considerably reduced parameters, alleviating the reliance on heavy pretrained model dependencies; and
\lowercase\expandafter{\romannumeral3}) Learning continuous function rather than discrete representations, which provides greater flexibility in creating embeddings for various downstream tasks.

\section{Related works}
\subsection{Masked image modelling}
Masked image modelling (MIM) has emerged as a promising approach in the field of self-supervised learning, enabling the derivation of robust representations by reconstructing occluded or masked imagery \cite{baevski2022data2vec,bao2021beit,dong2022bootstrapped,he2022masked,xie2022simmim,wei2022masked}.
The fundamental idea behind MIM is to artificially introduce occlusions in input data, followed by training a neural network to restore the original images from these masked versions.
This process encourages the model to extract and focus on meaningful features, thus yielding a more robust and informative representation \cite{pan2022towards}.

MIM can be categorized into two primary frameworks: the teacher-student framework and the MAE framework.
In the former, a pretrained "teacher" network guides a "student" network to restore occluded data \cite{baevski2023efficient,li2021mst,wei2022masked,wu2022extreme,zhou2022mimco}.
BEiT exemplifies this, considering the pretrained encoder as a fixed teacher and incorporating an additional layer mapping the path token to discrete pseudo labels \cite{bao2021beit}.
Conversely, the MAE framework leverages an encoder-decoder architecture, directly predicting the obscured regions \cite{he2022masked}.

Recent advancements in MIM are geared towards the convergence of both frameworks \cite{bai2023masked,lee2022exploring,yao2023moma}, refinement of cornerstone models like BEiT and MAE \cite{chen2022sdae,huang2022green,peng2022beit,woo2023convnext,wu2022denoising,xie2022masked}, and enhancing masking techniques \cite{chen2023improving,kakogeorgiou2022hide,li2022semmae,shi2022adversarial,wang2023hard}.
In contrast, limited work has been done on adapting MIM to different architectures, such as MaskClip and A$^2$MIM \cite{dong2023maskclip,li2022architecture}.

\subsection{Implicit neural representations}
Implicit neural representations (INRs) offer a promising alternative to traditional explicit representations, offering an innovative approach to depict complex geometries and continuous data without explicitly defining the underlying function.
Rather than directly storing pixel values or clear-cut geometric data, INRs represent the underlying scene implicitly as a continuous function, usually represented by a deep network like a coordinate-based multi-layer perceptron (MLP).
This function can project any pixel location to its associated properties without direct geometric storage \cite{chen2022transformers,dupont2022data,kim2023generalizable,tancik2021learned}.

The intrinsic nature of INRs provides a versatile framework, accommodating various input sizes and formats.
Prominent developments in INRs research include neural radiance fields (NeRF), which model a volumetric scene's radiance from sparse 2D observations as a function of 3D coordinate function \cite{mildenhall2021nerf}.
Additionally, the concept of INRs has been adapted for 2D image representation, enabling decoding for arbitrary output resolution \cite{anokhin2021image,chen2021learning}.

\section{Method}
In this section, we present the \emph{masked implicit neural representations (MINR)} framework, designed for masked interpolation of input samples using INRs.
Our method efficiently interpolates masked regions and offers robustness in out-of-distribution (OOD) settings.
We use the TransINR and GINR architecture as the backbone of our approach, enabling seamless generalization across diverse dataset instances \cite{chen2022transformers,kim2023generalizable}.
The overview of proposed framework is visualized in Figure \ref{fig:framework overview}.

\subsection{Integrating INRs with MIM}
Given a dataset $\mathcal{O}=\{o^{(n)}\}_{n=1}^N$ containing $N$ observations $o^{(n)}=(x_i^{(n)}, y_i^{(n)})$, we introduce random masks to produce the masked dataset $\mathcal{M}$:
\begin{equation}\label{eq: Masked modelling}
    \mathcal{M}=\{m^{(n)} \mid m^{(n)}\coloneqq\mathrm{Mask}^{(n)}(o^{(n)})\}_{n=1}^N.
\end{equation}
The primary goal of INRs is  to estimate a continuous function $f_\theta:\mathbb{R}^2\rightarrow\mathbb{R}^3$, which maps input coordinates to corresponding properties.
Traditional INRs optimize an MLP for each instance, minimizing the L2 loss:
\begin{equation}\label{eq: INR l2 loss}
    \mathcal{L}_n(\theta^{(n)};m^{(n)}) = {\frac{1}{H\times W}} \sum_{i=1}^{H\times W} {\Vert y_i^{(n)}-f_{\theta^{(n)}}(x_i^{(n)})\Vert_2^2}.
\end{equation}
Our INR approach only leverages implicit information present in the masked images, ensuring adaptability across different masking strategies.

\subsection{Utilizing hypernetwork for generalizability}
For INRs to be effective in OOD settings, it's crucial to train the model on individual instances.
However, this basic approach is computationally challenging with large-scale datasets, and such independent training approach restricts the generalizability to unseen instances.
Hence, we utilizes transformer-based hypernetwork architectures to modulate the MLP weights efficiently:
\begin{equation}\label{eq: weight of MLP}
\theta = \{W_l\}_{l=1}^{L}\subset\mathbb{R}^{\mathrm{in}_l\times\mathrm{out}_l},
\end{equation}
with $W_l$ denoting the weight of the $l$-th layer of the $L$-layer MLP.

\paragraph{TransINR-based approach.}
In \cite{chen2022transformers}, the hypernetwork predicts the entire set of INR weights $\theta^{(n)}=\{W_l\}_{l=1}^{L}$ simultaneously using the encoded information from the masked image $m^{(n)}$.
Given the masked observation $\mathcal{M}$, generalized INR are optimized according to Equation \ref{eq: INR l2 loss} once the predicted weights have been modulated into:
\begin{equation}
\label{eq: TransINR loss}
    \mathcal{L}(\Theta;\mathcal{M}) = \frac{1}{N} \sum_{n=1}^{N}\mathcal{L}_n(\theta^{(n)};m^{(n)}),
\end{equation}
where $\Theta=\{\theta^{(n)}\}_{n=1}^N$ represents INR weights for entire instances in the dataset.
However, this approach lacks robustness across instances, as weights are independently constructed.

\paragraph{GINR-based approach.}
Empirically found in \cite{kim2023generalizable}, partitioning the MLP hidden layers into instance-specific and instance-agnostic layers is effective in learning commonalities across instances.
Using the most performant configuration, we denote the second layer as instance-specific, with the other layers being instance-agnostic.
The instance-agnostic parameters $\phi$ are shared across all instances:
\begin{equation}
\label{eq: GINR loss}
    \mathcal{L}(\Theta,\phi;\mathcal{M}) = \frac{1}{N} \sum_{n=1}^N{\mathcal{L}_n(\theta^{(n)},\phi;m^{(n)}}),
\end{equation}
This design enables the MLP to learn patterns both common across a dataset and unique per instance, making it ideal for OOD settings.

\section{Experiments}
\begin{table*}[ht]
    \begin{center}
        \begin{tabular}{|ll|c|c|c|r|}
            \hline
            \multicolumn{2}{|c|}{Method} & \texttt{CEL} & \texttt{IMG} & \texttt{IND} & \# Param. \\
            \hline\hline
            \multirow{2}{*}{MAE} & Large & 15.018 & 14.693 & 15.181 & 313.6M \\
            & Base & 15.401 & 14.452 & 14.370 & 106.2M \\
            \hline
            \multirow{2}{*}{MINR} & TransINR & \textbf{21.865} & 18.737 & 17.756 & 44.5M \\
            & GINR & 21.680 & \textbf{19.358} & \textbf{18.622} & \textbf{43.7M} \\
            \hline
        \end{tabular}
    \end{center}
    \captionsetup{width=1.8\columnwidth}
    \caption{
        \textbf{Comparison of PSNR performances in ID mask reconstruction.}
        Columns represent the \texttt{CelebA}, \texttt{Imagenette}, and \texttt{MIT Indoor67} datasets, respectively, with the last column indicating the number of parameters.
    }
    \label{table:id}
\end{table*}

\begin{table*}[ht]
    \begin{center}
        \begin{tabular}{|ll|cc|cc|cc|}
            \hline
            \multicolumn{2}{|c|}{\multirow{2}{*}{Method}} & \multicolumn{2}{c|}{\texttt{CEL} $\to$} & \multicolumn{2}{c|}{\texttt{IMG} $\to$} & \multicolumn{2}{c|}{\texttt{IND} $\to$} \\
            & & \texttt{IMG} & \texttt{IND} & \texttt{CEL} & \texttt{IND} & \texttt{CEL} & \texttt{IMG} \\
            \hline\hline
            \multirow{2}{*}{MAE} & Large & 14.262 & 14.300 & 14.853 & 14.779 & 14.858 & 14.949 \\
            & Base & 14.508 & 14.464 & 14.499 & 14.558 & 13.831 & 14.069 \\
            \hline
            \multirow{2}{*}{MINR} & TransINR & \textbf{18.058} & \textbf{17.361} & 19.929 & 17.920 & 18.992 & 18.103 \\
            & GINR & 18.041 & 17.336 & \textbf{19.994} & \textbf{18.045} & \textbf{19.509} & \textbf{18.573} \\
            \hline
        \end{tabular}
    \end{center}
    \captionsetup{width=1.8\columnwidth}
    \caption{
        \textbf{Comparison of PSNR performances in OOD mask reconstruction.}
        The arrow ($\to$) indicates the source to target domain transfer.
    }
    \label{table:ood}
    \vspace{-9pt}
\end{table*}

In this section, we assess the efficacy of our proposed MINR method against the MAE approach for mask reconstruction tasks.
We conduct evaluations under both in-domain (ID) and out-of-domain (OOD) settings.

\subsection{Datasets}
We employ three diverse datasets for our experiments, namely \texttt{CelebA} \cite{howard2020fastai}, \texttt{Imagenette} \cite{liu2015deep}, and \texttt{MIT Indoor67} \cite{quattoni2009recognizing}:
\begin{itemize}
    \item \texttt{CelebA}: A comprehensive facial dataset encompassing 202K images.
    \item \texttt{Imagenette}: A subset of ImageNet, containing 10 distinct classes with a total of 7K images.
    \item \texttt{MIT Indoor67}: Specifically curated for indoor scene recognition, this dataset houses 15K images.
\end{itemize}
Notably, these datasets were selected due to their varied nature, which allows us to evaluate the robustness of our method across different data distributions.

\subsection{Experimental setup}
For our experiments, we maintain a consistent input image resolution of $182\times182$.
We employ a 5-layer MLP to define $f_\theta$.
Images are segmented into patches of size $14\times14$, of which a majority, 75\%, are subsequently masked out at random.
The configuration of our transformer-based hypernetwork is in alignment with the vision transformer architecture detailed in \cite{dosovitskiy2020image}.

The peak signal-to-noise ratio (PSNR) serves as our primary evaluation metric, a standard choice for reconstruction tasks \cite{dong2015image}.
To ensure the reliability of our results, we maintain consistent experimental settings, using models and hyperparameters as per their official implementations.

For ID evaluations, the test data originates from the same distribution as the training set.
Conversely, the OOD setting involves evaluation using data from a different distribution, aiming to gauge the model's generalization capability.

\subsection{Results}
The results of ID and OOD mask reconstruction experiments are summarized in Table \ref{table:id} and \ref{table:ood}, respectively.
Our results highlight that MINR consistently outperforms in both settings, achieving superior mask reconstruction with fewer parameters.
For the CelebA dataset, commonly used in mask reconstruction evaluations, there was about 6.4dB improvement in ID despite having more than half the parameters reduced.
Also, when evaluated on different data distribution from the training set, there was an improvement of more than 3dB for most cases.
This is further depicted in Figure \ref{fig:result_figure}, where MINR exhibits enhanced clarity in reconstructing masked patches.
Considering that MAE computes the loss exclusively on masked patches, we visualize the results with the pasted unmasked patches onto the reconstruction results for fair comparison.

\section{Conclusion}
In this work, we propose the MINR framework, a method that synergistically combines the principles of MIM with INRs to robustly tackle mask reconstruction tasks.
MINR leverages the continuous functional approximation capacity of INRs to improve both ID and OOD performance.
Our experimental evaluations against existing MAE approaches demonstrated the superiority of MINR in terms of reconstruction quality and robustness to diverse masking strategies, as substantiated by higher PSNR values across different datasets.
Moreover, our proposed framework significantly reduces the model parameters, thereby alleviating the need for heavy pretrained dependencies.
Finally, the adaptability of MINR's continuous function provides a flexible pathway for deriving feature embeddings across various downstream tasks.
In the future, we plan to leverage the flexibility of MINR, derived from its ability to learn a continuous function and agnosticism to input image sizes, to showcase its performance and applicability across various downstream tasks.

\ificcvfinal
\section{Acknowledgement}
Myungjoo Kang was supported by the NRF grant [2012R1A2C3010887] and the MSIT/IITP ([1711117093], [NO.2021-0-00077], [No. 2021-0-01343, Artificial Intelligence Graduate School Program (SNU)]).
\fi

\begin{small}
    \bibliographystyle{ieee_fullname}
    \bibliography{ref}
\end{small}

\end{document}